\documentclass[11pt,a4paper]{article}
\usepackage[hyperref]{acl2021}
\usepackage{latexsym}

\aclfinalcopy
\usepackage{microtype}

\usepackage{amsmath}
\usepackage{tabu}
\usepackage{svg}
\usepackage{booktabs}
\usepackage{multirow}
\usepackage{graphicx}
\usepackage{subfig}
\usepackage[T1]{fontenc}
\usepackage[T5]{fontenc}

\usepackage[utf8]{inputenc}

\usepackage{txfonts}

\newcommand{\consec}{\texttt{CONSEC}}
\newcommand{\rand}{\texttt{RAND}}
\newcommand{\randshift}{uniform-shift}
\newcommand{\trueshift}{sim-shift}
\newcommand{\enTOhe}{en\textrightarrow he}
\newcommand{\enTOvi}{en\textrightarrow vi}
\newcommand{\glTOen}{gl\textrightarrow en}
\newcommand{\skTOen}{sk\textrightarrow en}
\newcommand{\orig}{D_{\text{orig}}}
\newcommand{\new}{D_{\text{new}}}

\title{Data Augmentation by Concatenation for Low-Resource Translation: \\ A Mystery and a Solution}
\author{Toan Q. Nguyen \\
  University of Notre Dame \\
  \texttt{tnguye28@nd.edu} \And
  Kenton Murray \\
  Johns Hopkins University \\
  \texttt{kenton@jhu.edu} \\ \And
  David Chiang \\
  University of Notre Dame \\
  \texttt{dchiang@nd.edu} }

\begin{document}
\maketitle
\begin{abstract}
In this paper, we investigate the driving factors behind concatenation, a simple but effective data augmentation method for low-resource neural machine translation. Our experiments suggest that discourse context is unlikely the cause for concatenation improving BLEU by about +1 across four language pairs. Instead, we demonstrate that the improvement comes from three other factors unrelated to discourse: context diversity, length diversity, and (to a lesser extent) position shifting.
\end{abstract}

\section{Introduction}

Many attempts have been made to augment neural machine translation (MT) systems to use discourse context \citep{doc-level-1, doc-level-2, doc-level-3, doc-level-4, doc-level-5, doc-level-6, doc-level-7, doc-level-8, doc-level-9, context-1}. One particularly simple method is to concatenate consecutive pairs of sentence-pairs during training, but not during translation \cite{agrawal2018contextual, context-2, context-4, sent-concat}.\footnote{As this paper was being finalized, \citet{sent-concat} published independent work also presenting random concatenation as data augmentation for NMT. They find that concatenation helps the model translate long sentences better, while the focus of the present paper is to explain thoroughly why it helps.} In this paper, we confirm that this simple method helps, by roughly $+1$ BLEU across four low-resource language pairs. But we demonstrate that the reason it helps is \emph{not} discourse context, because concatenating \emph{random} pairs of sentence-pairs yields the same improvement.

Instead, we view concatenation as a kind of data augmentation or noising method (one which pleasantly requires no alteration to the text, unlike data augmentation methods that disturb word order \citep{nmt-noise-1, nmt-noise-2} or replace words with automatically-selected words \citep{nmt-aug-1, nmt-aug-2, nmt-aug-3}). 
Concatenating random sentences is easier than concatenating consecutive sentences, because many parallel corpora discard document boundaries, drop sentence-pairs, or even reorder sentence-pairs, so it can be difficult to know which sentence-pairs are truly consecutive. 

But the fact that random concatenation helps so much creates a mystery, which is the focus of the paper. If the reason is not discourse context, what is the reason?
We consider three new hypotheses:
\begin{itemize}
\item Random concatenation creates greater diversity of positions, because it lets the model see sentences shifted by effectively random distances.
\item Random concatenation creates greater diversity of contexts, helping the model learn what \emph{not} to attend to.
\item Random concatenation creates greater diversity of sentence lengths within a minibatch.
\end{itemize}
Through a careful ablation study, we demonstrate that all three of these factors more or less contribute to the improvement, and together completely explain the improvement.

\section{Concatenation}

We first present the concatenation methods and confirm that they improve low-resource translation.

\subsection{Methods} \label{sec:method}

Let $\orig = \{(x_i, y_i) \mid i=1, \ldots, N\}$ be the original training data. We consider two concatenation strategies:
\begin{description}
\item{\consec} Concatenate consecutive sentence-pairs: $\new = \{(x_i x_{i+1}, y_i y_{i+1}) \mid i=1, \ldots, N-1\}$. 
\item{\rand} Same as \consec, but randomly permute $\orig$ before concatenation.
\end{description}
For example, consider the following \enTOvi\ sentence pairs:
\begin{trivlist}
    \item \textit{And I think back .} $\rightarrow$ \textit{Và tôi nghĩ lại .}
    \item \textit{I think back to my father .} $\rightarrow$ \textit{Tôi nghĩ lại về cha tôi .}
\end{trivlist}
With \texttt{<BOS>}/\texttt{<EOS>} markings, the concatenated sentence-pairs would be:
\begin{trivlist}
    \item source input: \textit{And I think back . <EOS> I think back to my father . <EOS>}
    \item target input: \textit{<BOS> Và tôi nghĩ lại . <BOS> Tôi nghĩ lại về cha tôi .}
    \item target output: \textit{Và tôi nghĩ lại . <EOS> Tôi nghĩ lại về cha tôi . <EOS>}
\end{trivlist}

Since consecutive training examples often come from the same document, \consec\ lets the model look at some of the discourse context during training. In \rand, however, the concatenated sentences are almost always unrelated. In both cases, we train models on the combined data, $\orig \cup \new$.

\begin{table*}
\begin{minipage}{1.0\linewidth}
    \footnotesize
	\centering
	
	\begin{tabu}{@{}c|ccccccc@{}}
\toprule
      & train/dev/test sents. (x1000) & train steps/epoch & epochs & layers & heads & dropout & BPE ops. \\ \midrule
\textbf{\glTOen} & 10/0.68/1                     & 100                 & 1000          & 4         & 4        & 0.4     & 3k     \\
\textbf{\skTOen} & 61/2.27/2.45                     & 600                 & 200           & 6         & 8        & 0.3     & 8k     \\
\textbf{\enTOvi} & 133/1.55/1.27                     & 1500                & 200           & 6         & 8        & 0.3     & 8k     \\
\textbf{\enTOhe} & 210/4.52/5.51                     & 2000                & 200           & 6         & 8        & 0.3     & 8k     \\
\bottomrule
\end{tabu}
	
\end{minipage}
\caption{Some statistics of the datasets and models used.}
\label{tab:stats}
\end{table*}

% OVERALL RESULTS TABLE
\begin{table*}
\newcommand{\signif}{\rlap{$^\dagger$}}
\begin{minipage}{1.0\linewidth}
	\centering
	
	\begin{tabu}{@{}l|cccccccc|cccc@{}}
\toprule
      & \multicolumn{2}{c}{\glTOen} & \multicolumn{2}{c}{\skTOen} & \multicolumn{2}{c}{\enTOvi} & \multicolumn{2}{c|}{\enTOhe} & \multicolumn{4}{c}{average} \\
      & dev & test & dev & test & dev & test & dev & test & dev & $\Delta$ & test & $\Delta$ \\ \midrule
\textbf{baseline} & 22.9 & 20.7 & 29.2 & 30.3 & 29.0 & 32.7 & 30.3 & 28.1 & 27.8 &       & 28.0  \\
\textbf{\consec} & 24.9 & 22.9\signif & 30.3 & 31.5\signif & 29.2 & 33.5\signif & 30.6 & 28.6\signif & 28.8 & +1.0      & 29.1 & +1.1  \\
\textbf{\rand} & 25.3 & 23.1\signif & 30.3 & 31.6\signif  & 29.2 & 33.0 & 30.8 & 28.5\signif & 28.9 & +1.1       & 29.0 &  +1.0 \\
\bottomrule
\end{tabu}
	
\end{minipage}
\caption{Consecutive (\consec) and random (\rand) concatenation give the same BLEU improvement across our four low-resource language pairs. $\dagger$ = statistically significant improvement on the test set compared to baseline ($p < 0.01$).}
\label{tab:results}
\end{table*}

% COMPARE RAND vs CONSEC TABLE
\begin{table*}
\centering
\begin{tabular}{l|ccccc}
\toprule
&\multicolumn{5}{c}{dev BLEU} \\
 & \glTOen & \skTOen & \enTOvi & \enTOhe & avg    \\ 
\midrule
\textbf{\consec} & 23.5 & 29.6 & 29.7 & 31.1 & 28.5  \\ 
\textbf{\rand} & 24.0 & 29.2 & 29.4 & 31.3 & 28.5  \\
\bottomrule
\end{tabular}
\caption{Even when we concatenate consecutive sentence-pairs during translation, \consec\ does not outperform \rand. All BLEU scores in this table are computed on concatenated versions of the dev sets, and so are not comparable with the scores in other tables.}
\label{tab:abl-discourse}
\end{table*}

\subsection{Initial experiments}
\label{sec:setup}

We experiment on four low-resource language pairs: \{Galician, Slovak\} to English and English to \{Hebrew, Vietnamese\} \cite{data-src-1, data-src-2} using Transformer \cite{transformer}. We use the same setup as \citet{xmer-no-tears}, with PreNorm, FixNorm and ScaleNorm, as it has been shown to perform well on low-resource tasks. Since the data comes pre-tokenized, we only apply BPE. Data statistics and hyper-parameters are summarized in Table \ref{tab:stats}. 

For baseline, the training data is $\orig$. For concatenation, we first create $\new$, then combine it with $\orig$ to create the training data. Following \citet{batch-gen}, we randomly shuffle the training data and read it in chunks of 10k examples. Each chunk is sorted by source length before being packed into minibatches of roughly 4096 source/target tokens each.  

We calculate tokenized BLEU using \texttt{multi-bleu.perl} \cite{koehn-etal-2007-moses} and measure statistical significance using bootstrap resampling \cite{koehn-2004-statistical}. 

As seen in Table \ref{tab:results}, concatenation consistently outperforms the baseline across all datasets with significant improvement ($p < 0.01$) on almost every case. We observe that there is generally more improvement with less training data. For example, \enTOhe\ with more than 200k training examples gets only +0.5 BLEU, but \glTOen\ with only 10k sentences achieves +1.3 BLEU. On average, this method yields +1 BLEU over all four language pairs. We can also see that concatenating consecutive or random sentence pairs results in similar performance. For this reason, all the following ablation studies are conducted with \rand\ unless noted otherwise.

\section{Analysis}

Why does a method as simple as concatenation help so much? We reject the initial hypothesis that the model is assisted by discourse context (\S\ref{sect:discourse}) and consider three new hypotheses related to data augmentation (\S\ref{sec:position}--\S\ref{sec:length}).

\subsection{Discourse context}
\label{sect:discourse}

Since consecutive sentences often come from the same document, \consec\ provides the model with more discourse context during training. For \rand, however, the two sentences in a generated example are unlikely to have any relation at all. Despite this difference, we can see from Table \ref{tab:results} that both \consec\ and \rand\ achieve similar performance.  

To better understand whether discourse context plays any role here, we conduct a simple experiment. We perform concatenation just as in \consec\ and \rand, but on the dev set (as well as the training set), and measure BLEU on the concatenated dev set. The new BLEU scores are shown in Table \ref{tab:abl-discourse}, showing that even having discourse context available at translation time does not enable \consec\ to do better than \rand. While we acknowledge that there could be improvement due to discourse context that is not captured by BLEU, we can also say that the gain in BLEU that we do observe with concatenation is independent of the availability of discourse context.

\subsection{Position shifting}
\label{sec:position}

% PSEUDO-lENGTH TABLE
\begin{table*}
\begin{minipage}{1.0\linewidth}
	\centering
	
	\begin{tabu}{@{}c|l|cccccc@{}}
\toprule
      Row & & \glTOen & \skTOen & \enTOvi & \enTOhe & avg & $\Delta$ \\ \midrule
      
1 & \textbf{baseline} & 22.9 & 29.2 & 29.0 & 30.3 & 27.8  \\
2 & \textbf{baseline + \trueshift} & 22.7 & 29.8 & 29.0 & 30.4 & 28.0 & +0.2  \\
3 & \textbf{baseline + \randshift} & 23.8 & 29.8 & 29.3 & 30.5 & 28.4 & +0.6 \\
4 & \textbf{\rand} & 25.3 & 30.3 & 29.2 & 30.8 & 28.9 & +1.1 \\
5 & \textbf{\rand\ + \randshift} & 25.5 & 30.7 & 29.14 & 30.7 & 29.0 & +1.2 \\
\bottomrule
\end{tabu}
	
\end{minipage}
\caption{Position shifting improves accuracy somewhat, but the version of position shifting that mimics that of concatenation (\trueshift) gives less of an improvement than shifting by distances uniformly sampled from $[0, 100]$ (\randshift). All BLEU scores are on dev sets.}
\label{tab:abl-pos-shift}
\end{table*}

% DIVERSE CONTEXT
\begin{table*}
\begin{minipage}{1.0\linewidth}
	\centering
	
	\begin{tabu}{@{}c|l|cccccc@{}}
\toprule
      Row & & \glTOen & \skTOen & \enTOvi & \enTOhe & avg & $\Delta$ \\ \midrule
      
%0 & \textbf{baseline} & 22.86 & 29.15 & 29.01 & 30.26 & 27.82  \\
1 & \textbf{\rand} & 25.3 & 30.3 & 29.2 & 30.8 & 28.9  \\
2 & \textbf{\rand\ + mask} & 24.3 & 30.0 & 28.9 & 30.6 & 28.5 & $-0.4$  \\
3 & \textbf{\rand\ + sep-batch} & 24.9 & 30.1 & 29.1 & 30.6 & 28.7 & $-0.2$  \\
4 & \textbf{\rand\ + mask + sep-batch} & 23.2 & 29.8 & 29.3 & 30.5 & 28.2 & $-0.7$  \\
5 & \textbf{\rand\ + mask + sep-batch + reset-pos} & 23.1 & 29.6 & 28.9 & 30.5 & 28.0 & $-0.9$ \\
\bottomrule
\end{tabu}
\end{minipage}
\caption{Masking attention to prevent concatenated sentences from attending to one another (\textbf{mask}) reduces accuracy. Forming minibatches so as to prevent concatenation from increasing length diversity (\textbf{sep-batch}) also reduces accuracy. When we do both and also remove the effect of position shifting (\textbf{reset-pos}), we eliminate essentially all the improvement due to concatenation. All BLEU scores are on dev sets.}
\label{tab:abl-disverse}
\end{table*}

Since the Transformer uses absolute positional encodings, if a word is observed only a few times, the model may have difficulty generalizing to occurrences in other positions. Moreover, if there are too few long sentences, the model may have difficulty translating words very far from the start of the sentence. In concatenation, the second sentence is shifted by a random distance $n$ with $n$ being the first sentence's length in the sense that its positions are indexed from $n$ instead of $0$. We hypothesize that this allows the model to see, and thus, to be better-trained on more positions.     

If the improvement indeed comes from position shifting, we should be able to reproduce it without concatenation. In concatenation, we train on $\orig \cup \new$. While $\new$ has the same number of sentences as $\orig$ (\S\ref{sec:method}), each sentence is a concatenation of two sentences in $\orig$. This means that in total, 1/3 of sentences are shifted. So, we simulate the position-shifting that occurs in concatenation as follows. For each sentence-pair $(f_i, e_i)$ in the training data, with probability 1/3, choose a random training sentence pair $(f_j, e_j)$ and shift $f_i$ by $|f_j|$ and $e_i$ by $|e_j|$. We call this system \trueshift. 

%Furthermore, because $|f_j|$ and $|e_j|$ could be biased to a certain length bucket, 
We also try a more uniform shifting method, called \randshift, in which we sample, with probability 0.1, distances $s$ and $t$ uniformly from $[0, 100]$ and shift $f_i$ by $s$ and shift $e_i$ by $t$.  

\begin{figure}
\centering
\includegraphics[width=0.45\textwidth]{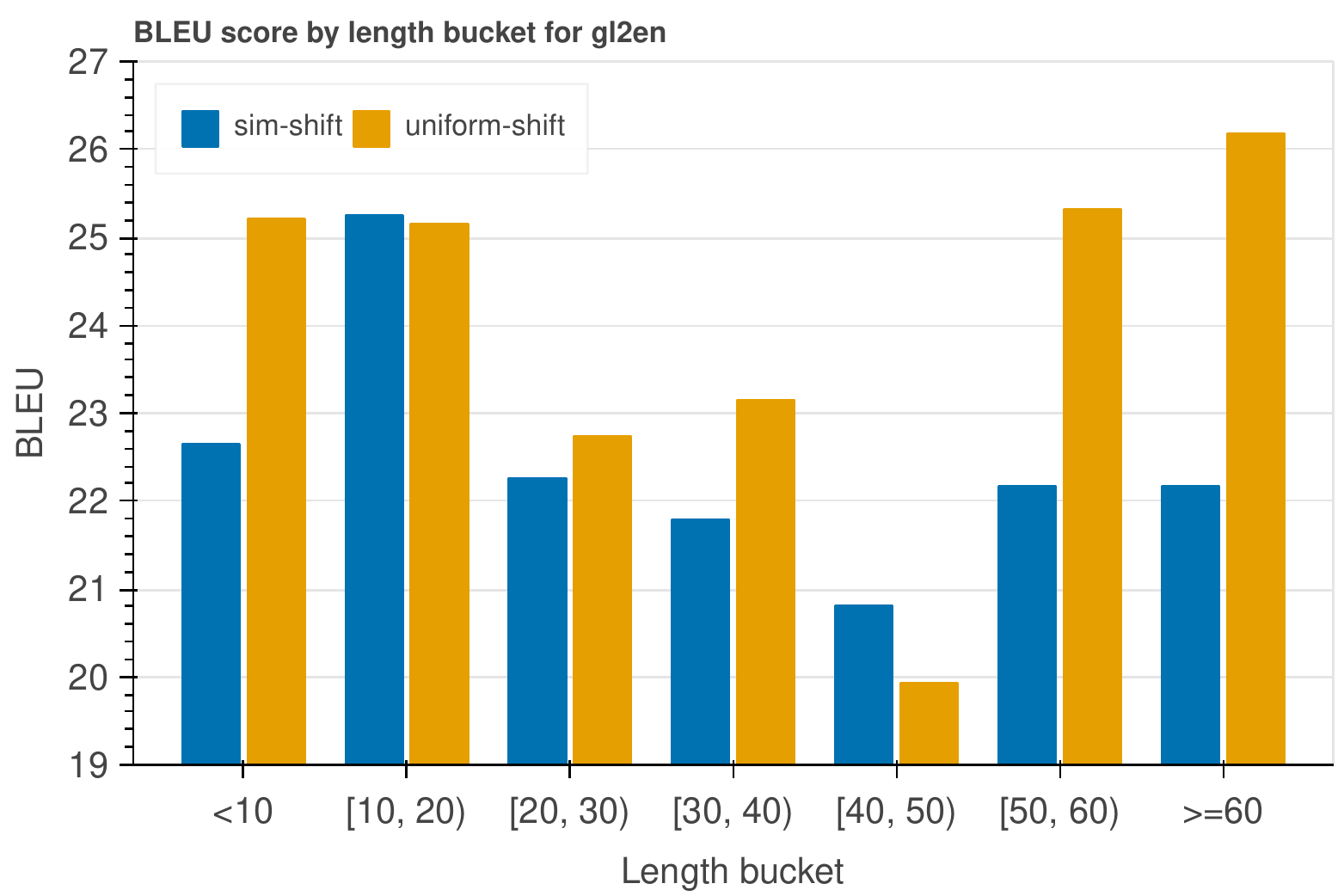}
\includegraphics[width=0.45\textwidth]{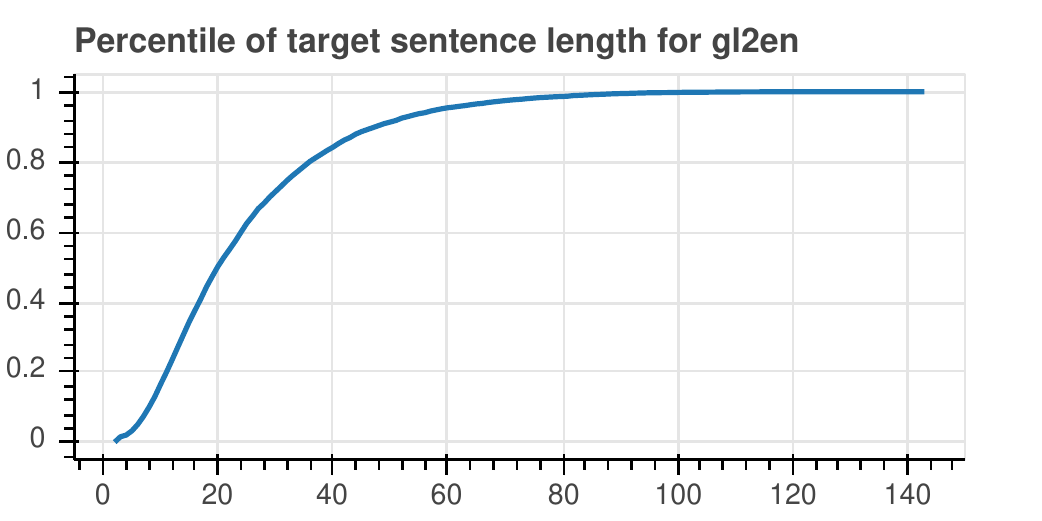}
\caption{gl2en: dev BLEU scores by length bucket (top) and its train length percentile (bottom).}
\label{fig:abs-pos-gl2en}
\end{figure}
Lines 1--3 in Table \ref{tab:abl-pos-shift} show that both \randshift\ and \trueshift\ do help somewhat. Surprisingly, \trueshift\ is outperformed by \randshift, especially for \glTOen\ with a gap of 0.9 BLEU. We attribute this to the fact that \randshift\ tends to shift sentences for longer distances and hence better generalizes to longer sentences. Indeed, as shown in Figure \ref{fig:abs-pos-gl2en} (bottom), most training sentences in \glTOen\ are shorter than 60. In Figure~\ref{fig:abs-pos-gl2en} (top), we see that \randshift\ outperforms \trueshift\ by the largest margin on the longest sentences. Nevertheless, adding \randshift\ on top of \rand\ (Table \ref{tab:abl-pos-shift}, row 5) only improves it very slightly.

To conclude, we show that position shifting can have a positive impact on low-resource NMT. However, it seems to contribute only a small part of the improvement due to concatenation, as we will confirm below (\S\ref{sec:ablation}).

\subsection{Context diversity}

In an attention layer, each query word is free to attend to any key word, and the model must learn to distinguish the keys that are related to a query from those that are not. Let us call the former \textit{positive contexts} and the latter \textit{negative contexts}. While positive contexts are important for determining how to translate a word, it is not trivial to generate more positive contexts, as it requires creating more parallel sentences that actually use the word. By contrast, creating more negative contexts is easy; this is what concatenation does. So one hypothesis is that concatenation helps by creating more negative contexts to improve the model's ability to attend to positive contexts.

To test this, we modify \rand\ by masking all self-attentions so that, in each concatenated example, each sentence can only attend to itself and not the other sentence. Similarly, in cross-attention, each target sentence can only attend to its corresponding source sentence, not the other one.
Table~\ref{tab:abl-disverse}, row 2 shows that this masking removes a large part of the improvement due to concatenation, showing that the availability of negative contexts during training does help during translation.

\subsection{Length diversity} 
\label{sec:length}
The last possible effect of concatenation that we consider is also the most subtle. Following previous work \citep{batch-gen, fairseq}, we first sort sentences by length, then splitting into minibatches of a fixed number of tokens. This puts sentences of similar lengths into the same minibatch, which improves computation efficiency as there is less padding. However, as observed by \citet{batch-gen}, short and long sentences are qualitatively different, so creating a minibatch of only short sentences or only long sentences approximates the full gradient less well than a minibatch of random sentences would.

With random concatenation, we again put examples of similar lengths into the same minibatch, but each example may consist of two sentences of very different lengths. Thus, it improves diversity within a minibatch while retaining efficiency. We hypothesize that this greater length diversity is part of the reason concatenation helps.

To evaluate this hypothesis, we try a different batch generation strategy from the one described above in Section~\ref{sec:setup}. In this setup, called \textbf{sep-batch}, we make two changes. First, the creation of $\new$ comes after sorting by sentence length (but before division into minibatches), so that in $\new$, each example comes from two similar-length ones. Second, we create batches from $\orig$ and $\new$ separately so there is no mixture of short sentences in $\orig$ and long sentences in $\new$.

As we can see in Table \ref{tab:abl-disverse}, removing length diversity (\textbf{sep-batch}, row 3) causes a small negative impact of $-0.2$ BLEU. So length diversity may be a contributing factor to concatenation's improvement.

\subsection{Feature ablation}
\label{sec:ablation}

We have shown that all three hypotheses (position diversity, context diversity, and length diversity) seem to contribute to the BLEU improvement due to concatenation. To see whether these hypotheses exhaustively explain it, we test all three together. First, we apply \textbf{mask} and \textbf{sep-batch} together, resulting in a drop of $-0.7$ BLEU (Table~\ref{tab:abl-disverse}, row~4). 

Finally, to remove the effect of position shifting, we additionally reset the positions of the second sentence in every concatenated example so they start at 0 again (\textbf{reset-pos}). 
Applying this on top of \textbf{mask} and \textbf{sep-batch}, it brings about the largest drop of $-0.9$ BLEU compared to \rand, resulting in a final model that is very close to the baseline (28.0 vs.~27.8 in Table~\ref{tab:abl-pos-shift}, row 4). Indeed, this model is only significantly different from the baseline on \skTOen\ ($p < 0.01$).
We conclude that these three hypotheses completely account for the improvement due to concatenation.

\section{Conclusion}

Random concatenation is a simple and surprisingly effective data augmentation method for low-resource NMT. Although the improvement of +1 BLEU it yields seems mysterious at first, we have shown that it can be explained by the fact that concatenation increases positions, context, and length diversity. Of these three factors, context diversity seems to be the most important. 

\section*{Acknowledgements}

This paper is based upon work supported in part by the Office of the Director of National Intelligence (ODNI), Intelligence Advanced Research Projects Activity (IARPA), via contract \#FA8650-17-C-9116. The views and conclusions contained herein are those of the authors and should not be interpreted as necessarily representing the official policies, either expressed or implied, of ODNI, IARPA, or the U.S. Government. The U.S. Government is authorized to reproduce and distribute reprints for governmental purposes notwithstanding any copyright annotation therein.

\bibliographystyle{acl_natbib}
\bibliography{anthology,acl2021}

%\appendix

\end{document}